\def\year{2020}\relax
\documentclass[letterpaper]{article} % DO NOT CHANGE THIS
\usepackage{aaai20}  % DO NOT CHANGE THIS
\usepackage{times}  % DO NOT CHANGE THIS
\usepackage{helvet} % DO NOT CHANGE THIS
\usepackage{courier}  % DO NOT CHANGE THIS
\usepackage[hyphens]{url}  % DO NOT CHANGE THIS
\usepackage{graphicx} % DO NOT CHANGE THIS
\usepackage{graphicx}  % DO NOT CHANGE THIS
\usepackage{mathtools}
\usepackage{amssymb,amsmath}
\usepackage{multirow}

\title{Improving Knowledge-aware Dialogue Generation via \\Knowledge Base Question Answering}

\author{
Jian Wang\textsuperscript{\rm 1},
Junhao Liu\textsuperscript{\rm 2},
Wei Bi\textsuperscript{\rm 3},
Xiaojiang Liu\textsuperscript{\rm 3},
Kejing He\textsuperscript{\rm 1}\textsuperscript{$\ast$},
Ruifeng Xu\textsuperscript{\rm 4},
Min Yang\textsuperscript{\rm 2}\thanks{Kejing He and Min Yang are corresponding authors.  This work was conducted when Jian Wang was interning at SIAT, Chinese Academy of Sciences.} \\
\textsuperscript{\rm 1}South China University of Technology, Guangzhou, China \\
\textsuperscript{\rm 2}Shenzhen Institutes of Advanced Technology, Chinese Academy of Sciences, Shenzhen, China \\
\textsuperscript{\rm 3}Tencent AI Lab, Shenzhen, China \\
\textsuperscript{\rm 4}Harbin Institute of Technology (Shenzhen), China \\
cs\_wangjian@mail.scut.edu.cn, \{jh.liu, min.yang\}@siat.ac.cn, \{victoriabi, kieranliu\}@tencent.com \\
kjhe@scut.edu.cn, xuruifeng@hit.edu.cn
}

\begin{document}
\maketitle
\begin{abstract}
Neural network models usually suffer from the challenge of incorporating commonsense knowledge into the open-domain dialogue systems. In this paper, we propose a novel knowledge-aware dialogue generation model (called TransDG), which transfers question representation and knowledge matching abilities from knowledge base question answering (KBQA) task to facilitate the utterance understanding and factual knowledge selection for dialogue generation. 
In addition, we propose a response guiding attention and a multi-step decoding strategy to steer our model to focus on relevant features for response generation. Experiments on two benchmark datasets demonstrate that our model has robust superiority over compared methods in generating informative and fluent dialogues. Our code is available at https://github.com/siat-nlp/TransDG.
\end{abstract}

\section{Introduction}
Building a dialogue system that is capable of providing informative responses is a long-term goal of artificial intelligence (AI).
Recent advances on dialogue systems are overwhelmingly contributed by deep learning techniques (i.e., sequence-to-sequence model) \cite{sutskever2014sequence}, which have  taken the state-of-the-art of dialogue systems to a new level.
However, fully data-driven neural models \cite{serban2015building,li2016persona,li2016deep}
tend to generate responses that are conversationally appropriate but seldom include factual content.
Previous studies \cite{ghazvininejad2018knowledge,zhou2018commonsense} revealed that infusing commonsense knowledge into dialogue systems could enhance user satisfaction and contribute to highly versatile and applicable open-domain dialogue systems. 

\begin{figure}[ht!]
\centering
\includegraphics[width=0.95\columnwidth]{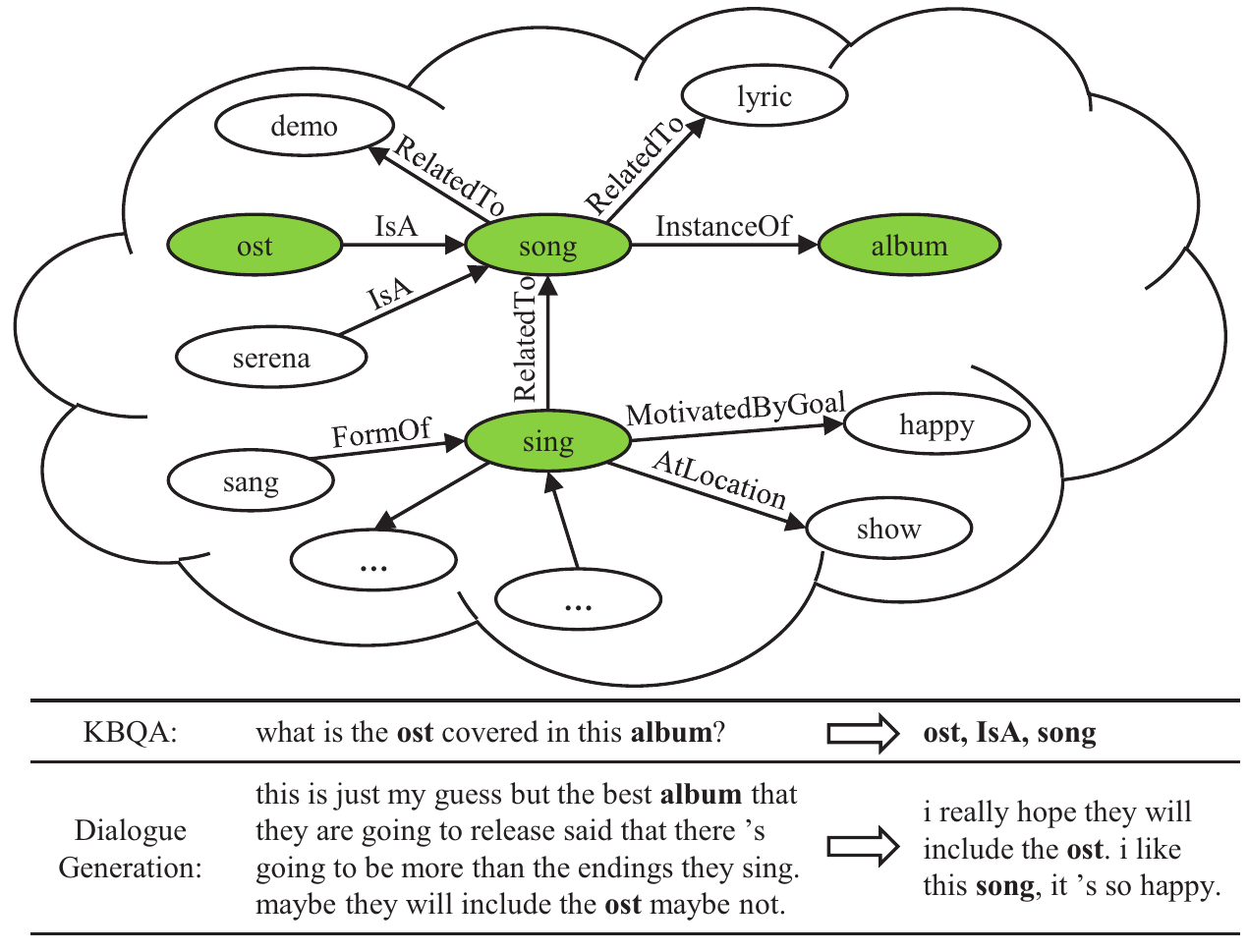}
\caption{Examples from a real-life dataset show that KBQA can facilitate the utterance understanding and factual knowledge selection for generating informative dialogue, e.g., locating the KB fact $\langle$ost, IsA, song$\rangle$.}
\label{fig:example}
\end{figure}

Several studies have been proposed to integrate external knowledge into dialogue generation \cite{zhu2017flexible,liu2018knowledge}. Despite the effectiveness of previous studies, there are still several challenges for generating informative and appropriate conversation, which are not addressed well in prior works. First, prior methods \cite{zhou2018commonsense} extract knowledge from KB by using each word in the post as query to retrieve related facts in an explicit manner. However, in dialogue systems, matching posts to exact facts in KB is much harder than explicit factoid inquiries answering. For some posts, the subjects and relations are elusive, e.g., the related entities are far from each other in the post (see Figure 1), which leads to trouble in matching relational facts from KB.  
Second, the generation-based methods generate the response word by word and lack a global perspective. As a result, the knowledge connection between the post and potential response (entity diffusion) is ignored, making the generated knowledge (entities) in responses not appropriate and reasonable with respect to the post. 
Third, most previous studies focus on enriching entities or triples for generation by merely incorporating information from KB. However, it is difficult to retrieve related facts and generate meaningful responses relying on solely the insufficient input posts, especially when the input posts are really short. 

To deal with the aforementioned challenges, we propose a novel knowledge-aware dialogue generation model (called TransDG), which effectively fuses external knowledge in KB into sequence-to-sequence model to generate informative dialogues by transferring the question modeling and knowledge matching abilities from KBQA, with the intuition that KBQA can facilitate the utterance understanding and factual knowledge (facts in KB) selection in dialogue generation (see Figure \ref{fig:example}). 
\textbf{First}, we pre-train a KBQA model, which consists of an encoding layer for representing both questions and candidate answers, and a knowledge selection layer for selecting the most appropriate answer from KB. \textbf{Second}, we encode the input post using gated recurrent unit (GRU) with augmentation of question representation layer learned from the KBQA model as dialogue encoder. 
\textbf{Third}, we integrate commonsense knowledge to generate informative responses by transferring the knowledge selection layer learned from the KBQA model with multi-step decoding strategy. The first-step decoder generates a draft response by attending relevant facts (entities) related to the post,  while the second-step decoder generates the  final response by referring to the context knowledge learned by the first-step decoder and improves the overall correctness of the generated response.
To further improve the informativeness of dialogues, we propose a response guiding attention mechanism, which leverages top-$k$ retrieved responses of similar posts to distill information in the input post.

\begin{figure*}[ht!]
\centering
\includegraphics[width=0.80\textwidth]{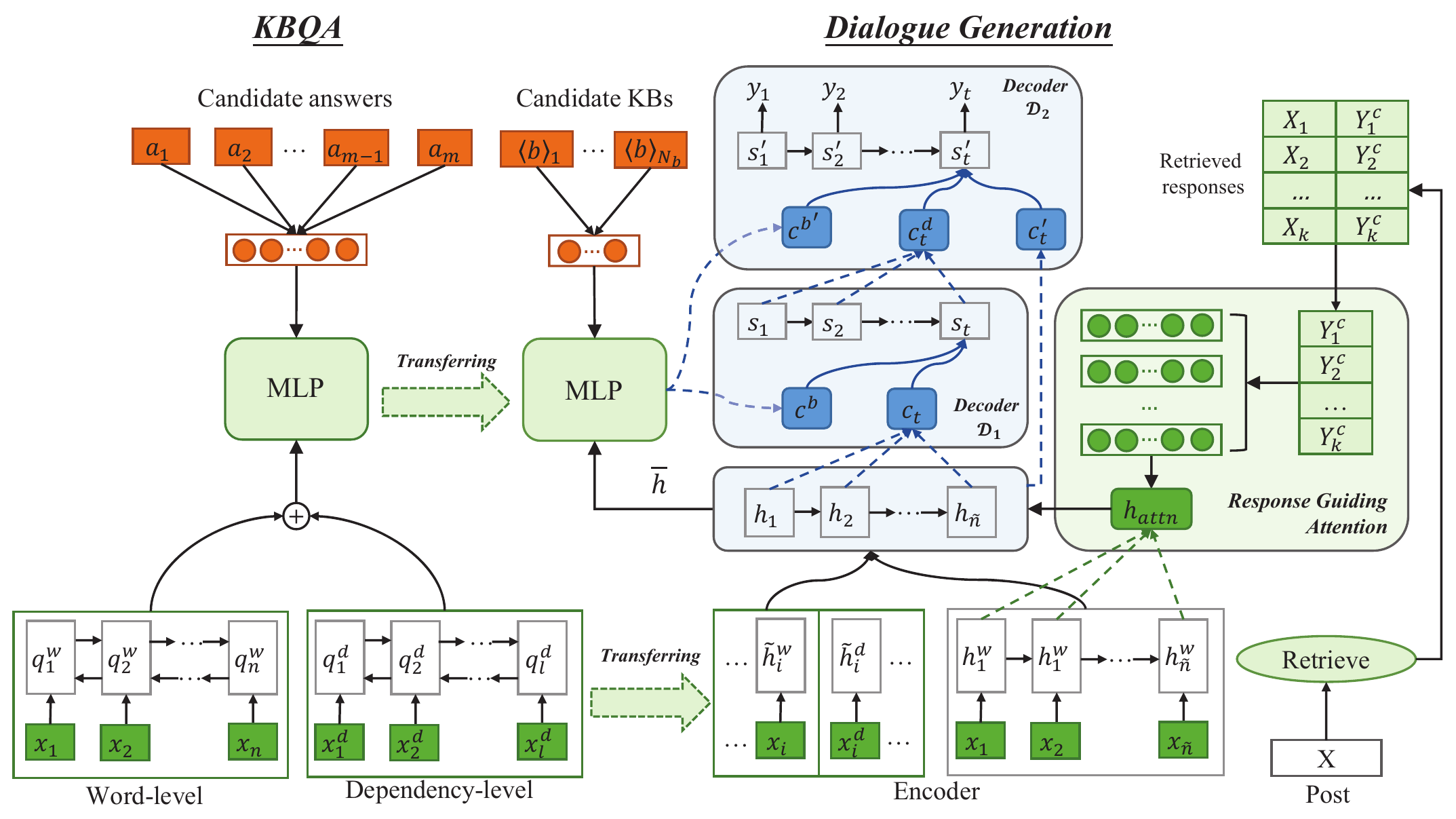}
\caption{Overview of TransDG model, which consists of a knowledge base question answering (KBQA) part (left) and knowledge-aware dialogue generation part (right), where the KBQA is pre-trained for transferring knowledge.}
\label{fig:figure2}
\end{figure*}

Our contributions are summarized as follows: 
\begin{itemize}
\item We propose TransDG, a novel knowledge-aware dialogue generation model, which transfers the abilities of question understanding and fact extraction from the pre-trained KBQA model to facilitate both post understanding and factual  knowledge selection from KB. 
\item We propose a multi-step decoding strategy which captures the knowledge connection between the post and response. Both the post and draft response generated by the first-step decoder is matched with relevant facts from KB, which makes the final response generated by the second-step decoder more appropriate and reasonable with respect to the post.
  
\item We propose a response guiding attention mechanism  which steers the model to focus on relevant features with the help of $k$-best response candidates.
\item Extensive experiments on a real dialogue dataset show that our model outperforms the compared methods from both quantitative and qualitative perspectives.
\end{itemize}

\section{Related Work}
Open domain dialogue generation \cite{serban2015building} aims at generating meaningful and coherent dialogue responses given input dialogue history. It is an important but challenging task that has received much attention recently from the natural language processing (NLP) community. 
Various techniques have been proposed to improve the quality of generated responses from different perspectives, such as diversity promotion \cite{li2016diversity}, unknown words handling \cite{gu2016incorporating}, prototype editing \cite{wu2018response} and retrieval-based ensemble \cite{song2018ensemble}. These models are end-to-end trainable and have good language modeling ability.
However, a well-known problem of these methods is that they are prone to generate universal and even meaningless responses.

Recently, incorporating external knowledge in open-domain dialogue generation is demonstrated to be effective to improve the performance of dialogue models. Some previous studies treated unstructured texts as external knowledge, which applied a convolutional neural network \cite{long2017knowledge} or a memory network \cite{ghazvininejad2018knowledge} to extract external knowledge for improving response generation. Many recent work incorporated open-domain knowledge bases in dialogue generation \cite{zhu2017flexible,liu2018knowledge,zhou2018commonsense,lian2019learning}. Specifically, related knowledge was acquired from knowledge bases to build knowledge grounded dialogues by using a copy network \cite{zhu2017flexible}. A neural knowledge diffusion model \cite{liu2018knowledge} was proposed to further integrate knowledge bases with dialogues through facts matching and entity diffusion. In addition, large scale commonsense knowledge bases were utilized in dialogue generation through graph attention mechanism \cite{zhou2018commonsense}. The posterior knowledge distribution was also utilized  to guide knowledge selection for response generation \cite{lian2019learning}. 

On the other hand, Knowledge Base Question Answering (KBQA) has also been an active research field in recent years, which aims at selecting an appropriate factual answer from structured knowledge bases, such as DBpedia \cite{auer2007dbpedia} and Freebase \cite{bollacker2008freebase},  given a query. 
A variety of methods have been proposed for KBQA, including information retrieval based methods \cite{yao2014information,xu2016question}, semantic parsing based methods \cite{yih2015semantic,hu2018answering} and neural network based methods \cite{hao2017end,yu2017improved,luo2018knowledge}. In most neural network based approaches, both questions and candidate answers (or knowledge facts) are encoded into distributed representations, and then the similarity calculation is used to select the most appropriate answer. For example, \citeauthor{luo2018knowledge} \shortcite{luo2018knowledge} employed an ensemble method to handle complex questions by leveraging dependency parsing and entity linking to enrich question representation \cite{luo2018knowledge}.

\section{Knowledge Base Question Answering}
As shown in Figure \ref{fig:figure2}, our model contains two parts: a KBQA model and a dialogue generation model, where knowledge learned from the KBQA task is transferred to dialogue generation in both encoding and decoding phases. In this section, we describe the KBQA model in detail. 

Given a question $Q=\{x_1,\ldots, x_{n}\}$, the task of KBQA is to select an appropriate answer from a set of candidate answers (facts) $A=\{a_1,\ldots,a_{m}\}$ in structured knowledge bases, where $n$ and $m$ are the lengths of the given question and candidate answer set, respectively. A common idea of KBQA is to encode the questions and facts in KBs into distributed representations, and then perform semantic matching between the question and candidate answer representations to obtain the final answer. 

\subsection{Encoding Layer}
\paragraph{Question Representation} 
We leverage both word-level and dependency-level information to learn the representation of question $Q$.  
For the word-level information, we convert each word $x_i$ into a word vector $\mathbf{v}_i^w \in \mathbb{R}^d$ through a word embedding matrix $\mathbf{E}^w \in \mathbb{R}^{|V_w| \times d}$, where $|V_w|$ is vocabulary size and $d$ is the size of word embedding. Then, we employ a bidirectional gated recurrent unit (BiGRU) \cite{cho2014learning} to obtain the hidden states of words in the question. Formally, the hidden state $\mathbf{q}^w_i$ at time step $t$ can be updated as:
\begin{equation}
\label{enc_by_word}
    \mathbf{q}^w_i = \text{BiGRU}^{w} (\mathbf{v}^w_i, \mathbf{q}^w_{i-1})
\end{equation}

To better capture the long-range relations between the words in the question, we follow \cite{xu2016question} to use the dependency path as additional representation, concatenating the words and dependency labels with directions. For example, the dependency path of question ``Who is the best actor in the movie'' is \{who, $\overrightarrow{nsubj}$, actor, $\overrightarrow{prep}$, in, $\overrightarrow{pobj}$, $\left\langle E\right \rangle$\}, where $\overrightarrow{nsubj}$, $\overrightarrow{prep}$ and $\overrightarrow{pobj}$ denote noun subjective, preposition and predicate objective respectively. $\left\langle E\right \rangle$ is a dummy token representing an entity word so as to learn a more general relation representation in syntactic level. We use the widely-used dependency parser Stanford CoreNLP \cite{manning2014stanford} to obtain dependency-level tokens, denoted as $x^d = \{x^d_1,\ldots,x^d_l\}$, where $l$ is the length of the dependency-level input, we convert each token $x_j^d$ into a word vector $\mathbf{v}_j^d \in \mathbb{R}^d$ through a  dependency embedding matrix $\mathbf{E}^d \in \mathbb{R}^{|V_d| \times d}$. Here, $\mathbf{E}^d$ is randomly initialized and updated during training, $|V_d|$ is the vocabulary size of the dependency tokens.
Then, we apply another BiGRU network to obtain the dependency-level question representation. Formally, the hidden state $\mathbf{q}^d_j$ is updated as:
\begin{equation}
\label{enc_by_dep}
    \mathbf{q}_j^{d}= \text{BiGRU}^d (\mathbf{v}_j^d, \mathbf{q}_{j-1}^d)
\end{equation}
 
We align the word-level and dependency-level sequences by padding, and combine their hidden states by element-wise addition:
$\mathbf{q}_i = \mathbf{q}_i^w + \mathbf{q}^d_i$. Hence, we can get the final question representation as $\mathbf{q} = [\mathbf{q}_1, \ldots, \mathbf{q}_n]$ . 

\paragraph{Candidate Answer Representation} Typically, the candidate answers in KBQA task are denoted as $A=\{a_1,\ldots,a_{m}\}$, where each answer $a_i$ is a fact from specific KB in the form of $\langle$subject entity, relation, object  entity$\rangle$. We encode such facts at both word-level and path-level. Given the word sequence of answers, we use the same word embedding matrix $\mathbf{E}^w$ to convert the words into word vectors $(\mathbf{a}_1^{w},\ldots,\mathbf{a}_m^{w})$, where $m$ is the length of the input answer. Then we calculate an average embedding as the word-level representation of the answer: $\mathbf{a}^{w}=\frac{1}{m}\sum_{i=1}^{m}\mathbf{a}_i^{w}$. For path-level, we treat each relation as a whole unit (e.g., ``related\_to'') and directly translate it into vector representation $\mathbf{a}^{p}$ through a KB embedding matrix $\mathbf{E}^{k} \in \mathbb{R}^{|V_k| \times d}$. Here, $\mathbf{E}^k$ is randomly initialized and updated during training, $|V_k|$ is the size of relations in the KB. The final representation of each candidate answer is defined as: $\mathbf{a} =\mathbf{a}^{w}+\mathbf{a}^{p}$.

\subsection{Semantic Matching and Model Training}

We calculate the semantic similarity score between question $q_i$ and candidate answer $a_j$ through a multi-layer perceptron: 
\begin{equation}
\label{mlp_score}
    S(q_i, a_j)=\text{MLP}([\mathbf{q}_i,\mathbf{a}_j]).
\end{equation}

%\subsection{Model Training}
During training, we adopt hinge loss to maximize the margin between positive answer set  and negative answer set:
\begin{equation}
    L_{q, A}=\max\{0, \gamma-S(q, A^{+})+S(q, A^{-})\}
\end{equation}
where $A^{+}$ are gold answers, $A^{-}$ are randomly sampled negative answers from the knowledge base for given questions, $\gamma$ is a parameter to tune margin between positive and negative samples.

\section{Knowledge-aware Dialogue Generation}
Given a post $X=\{x_1,\ldots, x_{\tilde{n}}\}$, the goal of dialogue generation is to generate a proper response $Y=\{y_1, \ldots, y_{\tilde{m}}\}$, where $\tilde{n}$ and $\tilde{m}$ are the lengths of the post and response, respectively. As shown in Figure \ref{fig:figure2}, our dialogue generation model transfers knowledge from KBQA task, facilitating the knowledge-level dialogue understanding and fact selection from KB. 

\subsection{Knowledge-aware Dialogue Encoder}
The dialogue generation employs a sequence-to-sequence (Seq2Seq) based method to generate a response for a given post. The encoder of Seq2Seq reads the post $X$ word by word and generates a hidden representation of each word by a GRU. Formally, given the input word embedding $\mathbf{e}_i^w$ for word $x_i$ in the post, the hidden state $\mathbf{h}_i^w$ can be updated by:
\begin{equation}
    \mathbf{h}_i^w= \text{GRU} (\mathbf{e}_i^w,\mathbf{h}_{i-1}^w)
\end{equation}

To facilitate the understanding of a post, we transfer the ability of question representation in KBQA task to obtain multi-level semantic understandings (i.e., word level and dependency level). Formally, we use the pre-trained bidirectional GRUs learned by KBQA task as additional encoders:
\begin{gather}
    \tilde{\mathbf{h}}_i^{w} =\text{BiGRU}^{w}(\mathbf{e}_i^{w}, \tilde{\mathbf{h}}_{i-1}^{w}) \\
    \tilde{\mathbf{h}}_i^{d} =\text{BiGRU}^{d}(\mathbf{e}_i^{d}, \tilde{\mathbf{h}}_{i-1}^{d})
\end{gather}
 We combine $\tilde{\mathbf{h}}^w_i$ and $\tilde{\mathbf{h}}^d_i$ by element-wise addition to obtain the KBQA based post representation: $\tilde{\mathbf{h}}_i = \tilde{\mathbf{h}}^w_i + \tilde{\mathbf{h}}^d_i$. 

\paragraph{Response Guiding Attention} To enrich the post representation for better comprehension, we propose a response guiding attention mechanism, which uses the retrieved responses of similar posts to steer the model to focus only on relevant information.
First, we use the widely-used text retrieval tool Lucene\footnote{https://lucene.apache.org/} to retrieve top-$k$ similar posts for a given post, the corresponding responses of the $k$ selected  posts serve as our candidate responses, denoted as $[Y_1^c, \ldots, Y_k^c]$. For each candidate response $Y_m^c$  ($m=1,\ldots,k$), we first convert it into word vectors $(\mathbf{y}_1,\ldots,\mathbf{y}_{\tilde{m}})$ through the word embedding matrix $\mathbf{E}^w$, and then represent the response by averaging operation: $e(Y_m^c)=\frac{1}{\tilde{m}}\sum_{j=1}^{\tilde{m}}\mathbf{y}_j$. We then calculate the mutual attention between the candidate response $Y_m^c$ and the hidden representation of the post to obtain the $m$-th candidate response guided post representation:
\begin{gather}
    \mathbf{h}_{attn}^{(m)}=\sum_{i=1}^{\tilde{n}}\alpha_{mi}\mathbf{h}_{i}, ~~ \alpha_{mi}=\text{softmax}(\beta_{mi}) \\
    \beta_{mi}=\mathbf{V}_h^T\tanh(\mathbf{W}_{h}e(Y_m^c)+\mathbf{U}_h\mathbf{h}_i)
\end{gather}
where $\mathbf{V}_h$, $\mathbf{W}_h$ and $\mathbf{U}_h$ are parameters to be learned. Finally, the candidate responses guided post representation is calculated by averaging the $k$ candidate responses guided post representations:
\begin{equation}
    \mathbf{h}_{attn}=\frac{1}{k}\sum_{m=1}^{k}\mathbf{h}_{attn}^{(m)}
\end{equation}
\noindent

Finally, the integrated post representation is formulated by concatenating $\mathbf{h}_{\tilde{n}}^w$, $\tilde{\mathbf{h}}_{\tilde{n}}$ and $\mathbf{h}_{attn}$, denoted as:
$\mathbf{h}_{final}=[\mathbf{h}_{\tilde{n}}^w;\tilde{\mathbf{h}}_{\tilde{n}}; \mathbf{h}_{attn}]$. 

\subsection{Knowledge-aware Multi-step Decoder}
The knowledge-aware decoder generates responses by transferring the knowledge selection ability learned from the pre-trained KBQA model using a multi-step decoding strategy. The first-step decoder generates a draft response by incorporating external knowledge relevant to the post. The second-step decoder generates the final response by referring to the post, the context knowledge and the draft response produced by the first-step decoder.  In this way, the multi-step decoder can capture the knowledge connection between the post and response, and therefore generate more coherent and informative response.

\paragraph{First-step decoder}
Formally, in the first-step decoder $\mathcal{D}_1$, the hidden state $\mathbf{s}_t$ of the decoder GRU at time step $t$ is updated as: 
\begin{equation}
    \mathbf{s}_t=\text{GRU}(\mathbf{s}_{t-1},  [\mathbf{c}_{t-1};\mathbf{c}^b;\mathbf{e}(\hat{y}_{t-1})])
\end{equation}
where $\mathbf{e}(\hat{y}_{t-1})$ is the embedding of previously generated word $\hat{y}_{t-1}$, $\mathbf{c}_{t-1}$ is the context vector at time step $t-1$, $\mathbf{c}^b$ is the attention vector augmenting the selected knowledge by KBQA model. Similar to previous studies, the context vector $\mathbf{c}_t$ is calculated as follows:
\begin{gather}
    \label{cal_context}
    \mathbf{c}_t=\sum_{k=1}^{\tilde{n}}\alpha_{tk}\mathbf{h}_k, ~~
    \alpha_{tk}=\text{softmax}(e_{tk})\\
    \label{cal_score}
    e_{tk}=\mathbf{V}_1^{T}\tanh(\mathbf{W}_1\mathbf{s}_{t-1}+ \mathbf{U}_1\mathbf{h}_k)
\end{gather}
where $\mathbf{V}_1$, $\mathbf{W}_1$ and $\mathbf{U}_1$ are parameters to be learned. 

The attention vector $\mathbf{c}^b$ is proposed to transfer the ability of selecting appropriate commonsense knowledge from the pre-trained KBQA model. Concretely, for a given post $X$, we first retrieve all the relevant triples from the knowledge base using the words in $X$ as queries, where each triple is represented as $\langle$subject entity, relation, object entity$\rangle$. All subject entities and object entities serve as our commonsense knowledge candidates, denoted as $\{\langle b_s,b_o\rangle_1,\ldots,\langle b_s,b_o\rangle_{N_b}\}$, where $b_s$ and $b_o$ represent subject and object entities, and $N_b$ is the number of the knowledge candidates. We take the encoded hidden representation of the post as input of the MLP layer defined in Eq. (\ref{mlp_score}) of the pre-trained KBQA model  to learn the correlation between the post and the candidate knowledge:
\begin{gather}
    \mathbf{c}^b=\frac{\sum_{j=1}^{N_b} r_{j}\mathbf{e}(b_j)}{\sum_{j=1}^{N_b}r_{j}}, ~~~~
    r_{j}=\text{MLP}([\overline{\mathbf{h}}, \mathbf{e}(b_j)])
\end{gather}
where $\mathbf{e}(b_j)$ is the concatenation of the word embeddings of the $j$-th subject and object entities, $\overline{\mathbf{h}}$ is the average value of the encoded hidden representation of the post, calculated as $\overline{\mathbf{h}}=\frac{1}{\tilde{n}}\sum_{k=1}^{\tilde{n}}\mathbf{h}_k$.

Therefore, the distribution of the draft response is given by:
\begin{equation}
     P(\hat{y}_t|Y_{<t},X)\propto \text{softmax}(\mathbf{W}_p[\mathbf{s}_t;\mathbf{c}_t;\mathbf{c}^b])
\end{equation}
where $\mathbf{W}_p$ is a trainable parameter.

\paragraph{Second-step decoder} For the second-step decoder $\mathcal{D}_2$, we take the hidden information generated by decoder $\mathcal{D}_1$ and the candidate knowledge into consideration. 
The second-step decoder can generate more appropriate  and reasonable responses with respect to the post by 
matching relevant facts from  KB  for  both  the  post  and  the  draft  response  generated  by  the  first-step  decoder. 
Formally, the hidden state of $\mathcal{D}_2$ is computed as:
\begin{equation}
    \mathbf{s}_{t}^{'}=\text{GRU}(\mathbf{s}_{t-1}^{'},[\mathbf{c}_{t-1}^{'};\mathbf{c}_{t-1}^d;\mathbf{c}^{b'};\mathbf{e}(y_{t-1})])
\end{equation}
where the computation of $\mathbf{c}_t^{'}$ is similar to that of $\mathbf{c}_t$ defined in Eq.(\ref{cal_context}) and Eq.(\ref{cal_score}). $\mathbf{c}_t^d$ is the first-step contextual information vector, which is defined as:
\begin{gather}
    \mathbf{c}_t^d=\sum_{k=1}^{T_s}\gamma_{tk}\mathbf{s}_k, ~~
    \gamma_{tk}=\text{softmax}(g_{tk})\\
    g_{tk}=\mathbf{V}_2^{T}\tanh(\mathbf{W}_2\mathbf{s}_{t-1}^{'}+ \mathbf{U}_2\mathbf{s}_k)
\end{gather}
where $\mathbf{s}_k$ is the hidden state of $k$-th time step in decoder $\mathcal{D}_1$, $T_s$ denotes the length of the time step in decoder $\mathcal{D}_1$.
The calculation of $\mathbf{c}^{b'}$ is similar to that of $\mathbf{c}^b$. The difference is that in decoder $\mathcal{D}_2$ we aim to capture the correlation between the draft response and the candidate knowledge:
\begin{gather}
    \mathbf{c}^{b'}=\frac{\sum_{j=1}^{N_b} r_{j}^{'}\mathbf{e}(b_j)}{\sum_{j=1}^{N_b}r_{j}^{'}}, ~~~~
    r_{j}^{'}=\text{MLP}([\overline{\mathbf{s}}, \mathbf{e}(b_j)])
\end{gather}
where $\overline{\mathbf{s}}$ is the average hidden representation in decoder $\mathcal{D}_1$, defined as: $\overline{\mathbf{s}}=\frac{1}{T_s}\sum_{k=1}^{T_s}\mathbf{s}_k$.

Finally, the generation distribution is formulated as follows:
\begin{equation}
    P(y_t|Y_{<t},X)\propto \text{softmax}(\mathbf{W}_o[\mathbf{s}_t^{'};\mathbf{c}_t^{'};\mathbf{c}_t^{d};\mathbf{c}^{b'}])
\end{equation}
where $\mathbf{W}_o$ is a trainable parameter.

\subsection{Model Training}
Our model is optimized in an end-to-end manner. We use $D$ to represent the training dataset, and $\Theta_e$, $\Theta_1$, $\Theta_2$ to represent the parameters of the encoder, the first-step decoder, the second-step decoder respectively. The training of the first-step decoding is to minimize the following loss:
\begin{equation}
\small
L(\mathcal{D}_1)=-\sum_{X,Y\in{D}}\sum_{t=1}^{\tilde{m}}\log P(\hat{y}_t|Y_{<t},X;\Theta_e, \Theta_1)
\end{equation}

Similarly, the second-step decoding is optimized by minimizing the following loss:
\begin{equation}
\small
L(\mathcal{D}_2)=-\sum_{X,Y\in{D}}\sum_{t=1}^{\tilde{m}}\log P(y_t|Y_<t,X;\Theta_e, \Theta_2)
\end{equation}

Finally, the total loss is the sum of $L(\mathcal{D}_1)$ and $L(\mathcal{D}_2)$:
\begin{equation}
\small
L = L(\mathcal{D}_1) + L(\mathcal{D}_2)
\end{equation}

\section{Experimental Setup}
\subsection{Datasets}
We use SimpleQuestions \cite{bordes2015large} dataset\footnote{http://fb.ai/babi} to train the KBQA model which consists of 75,910/10,845/21,687 instances for training/validation/testing, respectively. Each instance is a question paired with a knowledge triple as the gold answer. We use FB2M as the KB for SimpleQuestions, which is a subset of Freebase provided with the SimpleQuestions as default. It contains about 2M entities and 10M triples. Both the SimpleQuestions and the FB2M are used only for pre-training the KBQA model. The pre-trained KBQA model achieves 92.89 of F1 score on validation set.

For dialogue generation, we use Reddit \cite{zhou2018commonsense} single-round dialogue dataset\footnote{http://coai.cs.tsinghua.edu.cn/hml/dataset/\#commonsense}, which contains 3,384,185 training pairs, 10,000 validation pairs and 20,000 test pairs.
Each post-response pair is connected by one or more triple in ConceptNet\footnote{http://conceptnet.io}, which is used as the commonsense KB. 
It is noteworthy that we use the same dialogue dataset and commonsense knowledge base (i.e., ConceptNet) as in previous work \cite{zhou2018commonsense} for fair comparison. 
However, we would like to emphasize that the proposed model is general and can easily use other commonsense knowledge base (e.g., Freebase) to generate dialogues. The statistics of the datasets for dialogue generation and KBQA are shown in Table \ref{tab:table1}.

\begin{table}[htp]
\small
\centering
\begin{tabular}{c|c|c|c|c}
\hline
Task & \multicolumn{2}{c|}{QA / Dialog pairs} & \multicolumn{2}{c}{Knowledge Base} \\
\hline
\multirow{3}{*}{KBQA} & Training  & 75,910 & Entities & 2M   \\
\cline{2-5} & Validation & 10,845  & Relations & 31,940   \\ 
\cline{2-5} & Testing & 21,687 &  Triples   &  10M  \\ 
\hline
\multirow{3}{*}{Dialog} & Training  & 3,384,185 & Entities & 21,471 \\
\cline{2-5} & Validation & 10,000  & Relations & 44   \\ 
\cline{2-5} & Testing & 20,000 &  Triples &  120,850  \\ 
\hline
\end{tabular}
\caption{Statistics of datasets for different tasks.}
\label{tab:table1}
\end{table}

\subsection{Implementation Details}
For KBQA, we initialize word embeddings using Glove \cite{pennington2014glove} word vectors. The sizes of both word embedding and KB embedding  are set to 300. All BiGRUs are 1-layer BiGRU with 256 hidden units. The MLP model is a 2-layer fully-connected network with 512 hidden units and 1 hidden unit respectively. We set the margin $\gamma$ to be 0.5, and sample 20 negative samples for each gold answer. The model is trained using Adam \cite{kingma2014adam} optimizer with an initial learning rate 0.001. The batch size is set to 128. 

For dialogue generation, we also use 300 dimensional Glove word vectors to initialize the word embeddings. The vocabulary size is set to 30,000. The encoder and decoder have 2-layer GRUs with 512 hidden units for each layer. The dropout rate is set to 0.2. The number of candidate responses $k$ is set to 3, and we adopt the default configurations provided by the Lucene API. We train the model using Adam optimizer with an initial learning rate of 0.0005, and the batch size is set to 100.

\subsection{Baselines}
We compare our model with following suitable baselines:
\begin{itemize}
    \item \textbf{Seq2Seq:} a standard sequence-to-sequence model \cite{sutskever2014sequence}, which is widely used as a baseline in dialogue generation.
    \item \textbf{CopyNet:} a sequence-to-sequence based model with copy mechanism \cite{zhu2017flexible}, which acquires knowledge by copying entity words from related knowledge bases.
    \item \textbf{MemNet:} a knowledge-grounded model which uses memory units to process knowledge triples \cite{ghazvininejad2018knowledge}.
    \item \textbf{CCM:} a knowledge-aware dialogue generation model with static and dynamic graph attention mechanisms \cite{zhou2018commonsense}.
    \item \textbf{PostKS:} a knowledge-guided dialogue generation model which employs the posterior knowledge distribution to guide the knowledge selection \cite{lian2019learning}.
\end{itemize}

\subsection{Evaluation Metrics}
Both automatic and human evaluation metrics are used to measure the performance of our model. For automatic evaluation, we adopt \textbf{perplexity} and \textbf{entity score} \cite{zhou2018commonsense} as evaluation metrics, following previous work \cite{zhou2018commonsense}. The perplexity is widely used to quantify a language model, where a model performs better when perplexity is smaller. The entity score is used to measure the ability to generate relevant entities per response from the commonsense KB. Higher entity score generally indicates the generated response is more diverse. To further evaluate the quality of dialogue systems, we also adopt \textbf{BLEU} \cite{papineni2002bleu} as another automatic metric, which calculates $n$-gram overlaps between the generated response and the gold response.

We use human evaluation to evaluate the dialogue generation models from three perspectives: 
 \textbf{fluency}, \textbf{knowledge relevance} and \textbf{correctness} \cite{liu2018knowledge}. All the values are scored from 0 to 3, where higher score means better performance. Specifically, 500 posts are randomly selected from the test set, resulting in 3,000 responses generated by TransDG and baseline models in total for human evaluation. Three annotators are recruited to independently assign three scores for each generated response. 
 The agreement ratio computed with Fless' kappa \cite{fleiss1971measuring} is 0.58, showing moderate agreement.
 We report the average rating scores from all annotators as the final human evaluation results.

\section{Experimental Results}
\subsection{Quantitative Results}

\begin{table}[ht]
\small
\centering
\begin{tabular}{c|c|c|c|c|c}
\hline Model & Overall & High & Medium & Low & OOV \\
\hline Seq2Seq & 47.02 & 42.41 & 47.25 & 48.61 & 49.96 \\
MemNet & 46.85 & 41.93 & 47.32 & 48.86 & 49.52 \\ 
CopyNet & 40.27 & 36.26 & 40.99 & 42.09 & 42.24 \\
CCM & 39.18 & 35.36 & 39.64 & 40.67 & 40.87 \\
PostKS & 43.56 & 40.65  & 44.06  &  46.36 & 49.32  \\
\hline
TransDG & \textbf{37.53} & \textbf{32.18} & \textbf{36.12} & \textbf{38.46} & \textbf{40.75} \\
\hline
\end{tabular}
\caption{Automatic evaluation with \textbf{perplexity}.}
\label{tab:table2}
\end{table}

\begin{table}[ht]
\small
\centering
\begin{tabular}{c|c|c|c|c|c}
\hline Model & Overall & High & Medium & Low & OOV \\
\hline Seq2Seq & 0.717 & 0.713 & 0.740 & 0.721 & 0.669 \\
MemNet & 0.761 & 0.764 & 0.788 & 0.760 & 0.706 \\ 
CopyNet & 0.960 & 0.910 & 0.970 & 0.960 & 0.960 \\
CCM & 1.180 & 1.156 & 1.191 & 1.196 & 1.162 \\
PostKS & 1.041  & 1.007  & 1.028  & 0.993 & 0.978  \\
\hline
TransDG & \textbf{1.207} & \textbf{1.195} & \textbf{1.204} & \textbf{1.232} & \textbf{1.182} \\
\hline
\end{tabular}
\caption{Automatic evaluation with \textbf{entity score}.}
\label{tab:table3}
\end{table}

\begin{table}[ht]
\small
\centering
\begin{tabular}{c|c|c|c|c}
\hline Model & BLEU-1 & BLEU-2 & BLEU-3 & BLEU-4 \\
\hline Seq2Seq & 0.0977 & 0.0098 & 0.0012  & 0.0002  \\
MemNet &  0.1652 & 0.0174  & 0.0028  & 0.0004 \\ 
CopyNet & 0.1715  & \textbf{0.0181}  & 0.0029 & 0.0005  \\
CCM & 0.1625  & 0.0175  & 0.0030  & 0.0005 \\
PostKS & 0.1683  & 0.0165  & 0.0029  & 0.0004  \\
\hline
TransDG & \textbf{0.1807} & 0.0178 & \textbf{0.0031} & \textbf{0.0006} \\
\hline
\end{tabular}
\caption{Automatic evaluation with \textbf{BLEU}.}
\label{tab:table4}
\end{table}

As shown in Table \ref{tab:table2}, TransDG achieves the lowest perplexity on all the datasets, indicating that the generated responses are more grammatical. 
Table \ref{tab:table3} demonstrates that the models leveraging external knowledge achieve better performance than the standard Seq2Seq model in generating meaningful entity words and diverse responses. 
In particular, our model outperforms all the baselines significantly with highest entity score. This verifies the effectiveness of transferring knowledge from KBQA task for factual knowledge selection. 
The BLEU values shown in Table \ref{tab:table4} demonstrates the comparison results from word-level overlaps. TransDG tends to generate responses that are more similar to the gold responses than baselines in most cases. This may be because that our model utilizes retrieved candidate responses to provide guidance. In addition, we observe that CopyNet also performs well in terms of BLEU score, since it incorporates copying mechanism into the decoding process, which can copy words or subsequences from the input post and KB. 

The human evaluation results are reported in Table \ref{tab:table5}.  As shown in Table \ref{tab:table5}, TransDG tends to generate more appropriate and informative responses in terms of human annotation. Specifically, the responses generated by TransDG have higher knowledge relevance than other models, indicating that TransDG is effective to incorporate appropriate commonsense knowledge.

\begin{table}[ht!]
\small
\newcommand{\tabincell}[2]{\begin{tabular}{@{}#1@{}}#2\end{tabular}}
\centering
\begin{tabular}{c|c|c|c}
\hline
Model & Fluency & Relevance & Correctness \\
\hline
Seq2Seq & 1.67 & 0.68 & 0.80 \\
MemNet & 1.83 & 0.89 & 1.32 \\
CopyNet & 2.36 & 1.13 & 1.08 \\
CCM & 2.27 & 1.35 & 1.22 \\
PostKS & 2.32  & 1.36  & 1.31  \\
\hline
TransDG & \textbf{2.41} & \textbf{1.52} & \textbf{1.34} \\
\hline
\end{tabular}
\caption{Human evaluation result.}
\label{tab:table5}
\end{table}

\subsection{Case Study}
Table \ref{tab:table6} lists some responses generated by TransDG and the baselines. The Seq2Seq model is unable to comprehend the post since it does not incorporate the external commonsense knowledge, which further verifies the effectiveness of commonsense knowledge for open-domain dialogue generation. MemNet and CopyNet can generate fluent responses. However, these responses are not appropriate with respect to the conversation context. CCM and PostKS are capable of generating some informative words while the whole responses are lack of continuity in logic. Instead, the proposed TransDG can generate a more reasonable response with appropriate entity words such as ``fan'' and ``game''.

\begin{table}[ht!]
\small
\centering
\resizebox{0.95\columnwidth}{!}{
\begin{tabular}{p{7.0cm}} 
\hline
\textbf{Post:} Did you play Mario Party 2?\\
\textbf{Reference:} Yes! I played this game recently with my friends on the N64, and it was amazing. I forgot how much fun I had with this as a kid. \\
\textbf{Seq2Seq:} It's a game, but it's not.\\
\textbf{MemNet:} I'm just going to have to wait for the next party to be in the game.\\
\textbf{CopyNet:} I'm not sure I'd be happy with that party. \\
\textbf{CCM:} I'm not a big Mario Party, but I'm a little disappointed. \\
\textbf{PostKS}: I have not played Mario Party yet, but I have not played it yet. \\
\textbf{TransDG:} I'm a big fan of Mario Party, but I'm not sure if I can get a copy of the game or play it on my pc.\\
\hline
\end{tabular}}
\caption{Case study of generated responses. The reference means the ground-truth response in the dataset.}
\label{tab:table6}
\end{table}

\subsection{Ablation Study}
To investigate the effectiveness of each module proposed in our model, we conduct ablation test by removing the following modules: (1) the pre-trained question representation module transferred from KBQA (w/o QRT), which encodes the posts with a standard BiGRU defined by question encoding layer with random initialization, (2) the knowledge selection module transferred from KBQA (w/o KST), (3) the response guiding attention module (w/o RGA), and (4) the second-step decoder module (w/o SSD).
The ablation test results are reported in Table \ref{tab:ablation}. From the results, we can observe that the performance of TransDG drops sharply when we discard the question representation module and the knowledge selection module transferred from KBQA. 
This is within our expectation since  the question representation module transferred from KBQA helps the encoder to capture essential information (e.g., informative entities) from the post, and the knowledge selection module encourages the decoder to select appropriate facts from external KB. Response guiding attention also has noticeable impact on the performance of TransDG, especially on BLEU scores. The candidate responses highlight relevant context and suppress unimportant ones, enabling to generate more accurate responses. 
In addition, the second-step decoder can improve the ability of TransDG to generate relevant entities per response. It is no surprise that combining all the factors achieves the best performance for all evaluation metrics.

\begin{table}[ht!]
\small
\centering
\begin{tabular}{c|c|c|c|c}
\hline
Model &  Perplexity &  Entity & BLEU-1 & BLEU-2 \\
\hline
TransDG & 37.53  & 1.207 & 0.1807 & 0.0178 \\
 w/o QRT & 42.17 & 1.076 & 0.1604 & 0.0171 \\
 w/o KST & 43.05 & 0.774 & 0.1643 & 0.0158 \\
 w/o QRT+KST & 44.15 & 0.772 & 0.1612 & 0.0170 \\
 w/o RGA & 38.62 & 1.106 & 0.1712 & 0.0170 \\
 w/o SSD & 38.18 & 1.114 & 0.1804 & 0.0178 \\
\hline
\end{tabular}
\caption{Ablation results of TransDG on the test set. Here, Entity represents entity score.}
\label{tab:ablation}
\end{table}

\subsection{Error Analysis}
To better understand the limitations of TransDG, we additionally carry out an analysis of the errors made by TransDG. We randomly select 100  responses generated by TransDG that achieve low human evaluation scores. We reveal several reasons for low human evaluation scores which can be divided into the following categories. 

\textit{Illogical} (36\%): The top error category is illogical, including the responses that are contradictory or conflict with the input posts. For example, the response ``I'm not sure he's a good player, especially he's a good player'' has a score of 0 since it lacks continuity in logic.   This type of errors are difficult to handle with current techniques, especially when trying to build an end-to-end model.

\textit{Miscellaneous} (32\%):  The second most common error category is miscellaneous, which includes responses that are less informative or not reasonable caused by polysemic words. For example, given a post ``You live under a rock? That saying is older than dirt'',  the generated response is ``I'm not a fan of the rock music''. 

\textit{Irrelevant} (20\%): The third error category includes the responses that are fluent but not relevant to the post or too general to correctly answer the post, which occurs when the model fails to incorporate appropriate knowledge. For instance,
a general response ``I don't know, I'm glad I could build a good one!'' is generated for a help-seeking post ``Is there a tutorial to the design you used anywhere? Or maybe anything to help me build it?'' 

\textit{Ungrammatical} (12\%): Another error category includes the responses that are grammatically incorrect (e.g., ``it's not a single color color''). This may be because that the model fails to prevent the generation of repeated words. 

\section{Conclusion}
In this paper, we propose a novel knowledge-aware dialogue generation model TransDG, which is the first neural dialogue model that incorporates commonsense knowledge via transferring the abilities of utterance representation and knowledge selection from KBQA task. In addition, we propose a response guiding attention mechanism to enhance the input post understanding in encoder, and refine the knowledge selection by multi-step decoding to generate more appropriate and informative responses. Extensive experiments  demonstrate the effectiveness of our model.

\section*{Acknowledgments}
This work was partially supported by the National Natural Science Foundation of China (NSFC) (No. 61272200 and No. 61906185),  CCF-Tencent Open Research Fund, Science and Technology Planning Project of Guangdong Province (No. 2017B030306016), Guangdong Basic and Applied Basic Research Foundation (No. 2019A1515011705). 

\bibliography{AAAI-WangJ.4018}

\begin{thebibliography}{}

\bibitem[\protect\citeauthoryear{Auer \bgroup et al\mbox.\egroup
  }{2007}]{auer2007dbpedia}
Auer, S.; Bizer, C.; Kobilarov, G.; Lehmann, J.; Cyganiak, R.; and Ives, Z.
\newblock 2007.
\newblock Dbpedia: A nucleus for a web of open data.
\newblock In {\em The semantic web}. Springer.
\newblock  722--735.

\bibitem[\protect\citeauthoryear{Bollacker \bgroup et al\mbox.\egroup
  }{2008}]{bollacker2008freebase}
Bollacker, K.; Evans, C.; Paritosh, P.; Sturge, T.; and Taylor, J.
\newblock 2008.
\newblock Freebase: a collaboratively created graph database for structuring
  human knowledge.
\newblock In {\em SIGMOD},  1247--1250.
\newblock AcM.

\bibitem[\protect\citeauthoryear{Bordes \bgroup et al\mbox.\egroup
  }{2015}]{bordes2015large}
Bordes, A.; Usunier, N.; Chopra, S.; and Weston, J.
\newblock 2015.
\newblock Large-scale simple question answering with memory networks.
\newblock {\em arXiv preprint arXiv:1506.02075}.

\bibitem[\protect\citeauthoryear{Cho \bgroup et al\mbox.\egroup
  }{2014}]{cho2014learning}
Cho, K.; Van~Merri{\"e}nboer, B.; Gulcehre, C.; Bahdanau, D.; Bougares, F.;
  Schwenk, H.; and Bengio, Y.
\newblock 2014.
\newblock Learning phrase representations using rnn encoder-decoder for
  statistical machine translation.
\newblock In {\em EMNLP},  1724--1734.

\bibitem[\protect\citeauthoryear{Fleiss}{1971}]{fleiss1971measuring}
Fleiss, J.~L.
\newblock 1971.
\newblock Measuring nominal scale agreement among many raters.
\newblock {\em Psychological bulletin} 76(5):378.

\bibitem[\protect\citeauthoryear{Ghazvininejad \bgroup et al\mbox.\egroup
  }{2018}]{ghazvininejad2018knowledge}
Ghazvininejad, M.; Brockett, C.; Chang, M.-W.; Dolan, B.; Gao, J.; Yih, W.-t.;
  and Galley, M.
\newblock 2018.
\newblock A knowledge-grounded neural conversation model.
\newblock In {\em AAAI}.

\bibitem[\protect\citeauthoryear{Gu \bgroup et al\mbox.\egroup
  }{2016}]{gu2016incorporating}
Gu, J.; Lu, Z.; Li, H.; and Li, V.~O.
\newblock 2016.
\newblock Incorporating copying mechanism in sequence-to-sequence learning.
\newblock In {\em ACL},  1631--1640.

\bibitem[\protect\citeauthoryear{Hao \bgroup et al\mbox.\egroup
  }{2017}]{hao2017end}
Hao, Y.; Zhang, Y.; Liu, K.; He, S.; Liu, Z.; Wu, H.; and Zhao, J.
\newblock 2017.
\newblock An end-to-end model for question answering over knowledge base with
  cross-attention combining global knowledge.
\newblock In {\em ACL},  221--231.

\bibitem[\protect\citeauthoryear{Hu \bgroup et al\mbox.\egroup
  }{2018}]{hu2018answering}
Hu, S.; Zou, L.; Yu, J.~X.; Wang, H.; and Zhao, D.
\newblock 2018.
\newblock Answering natural language questions by subgraph matching over
  knowledge graphs.
\newblock {\em IEEE Transactions on Knowledge and Data Engineering}
  30(5):824--837.

\bibitem[\protect\citeauthoryear{Kingma and Ba}{2014}]{kingma2014adam}
Kingma, D.~P., and Ba, J.
\newblock 2014.
\newblock Adam: A method for stochastic optimization.
\newblock {\em arXiv preprint arXiv:1412.6980}.

\bibitem[\protect\citeauthoryear{Li \bgroup et al\mbox.\egroup
  }{2016a}]{li2016diversity}
Li, J.; Galley, M.; Brockett, C.; Gao, J.; and Dolan, B.
\newblock 2016a.
\newblock A diversity-promoting objective function for neural conversation
  models.
\newblock In {\em NAACL},  110--119.

\bibitem[\protect\citeauthoryear{Li \bgroup et al\mbox.\egroup
  }{2016b}]{li2016persona}
Li, J.; Galley, M.; Brockett, C.; Spithourakis, G.~P.; Gao, J.; and Dolan, B.
\newblock 2016b.
\newblock A persona-based neural conversation model.
\newblock In {\em ACL},  994--1003.

\bibitem[\protect\citeauthoryear{Li \bgroup et al\mbox.\egroup
  }{2016c}]{li2016deep}
Li, J.; Monroe, W.; Ritter, A.; Galley, M.; Gao, J.; and Jurafsky, D.
\newblock 2016c.
\newblock Deep reinforcement learning for dialogue generation.
\newblock In {\em EMNLP},  1192--1202.

\bibitem[\protect\citeauthoryear{Lian \bgroup et al\mbox.\egroup
  }{2019}]{lian2019learning}
Lian, R.; Xie, M.; Wang, F.; Peng, J.; and Wu, H.
\newblock 2019.
\newblock Learning to select knowledge for response generation in dialog
  systems.
\newblock In {\em IJCAI},  5081--5087.

\bibitem[\protect\citeauthoryear{Liu \bgroup et al\mbox.\egroup
  }{2018}]{liu2018knowledge}
Liu, S.; Chen, H.; Ren, Z.; Feng, Y.; Liu, Q.; and Yin, D.
\newblock 2018.
\newblock Knowledge diffusion for neural dialogue generation.
\newblock In {\em ACL},  1489--1498.

\bibitem[\protect\citeauthoryear{Long \bgroup et al\mbox.\egroup
  }{2017}]{long2017knowledge}
Long, Y.; Wang, J.; Xu, Z.; Wang, Z.; Wang, B.; and Wang, Z.
\newblock 2017.
\newblock A knowledge enhanced generative conversational service agent.
\newblock In {\em Proceedings of the 6th Dialog System Technology Challenges
  (DSTC6) Workshop}.

\bibitem[\protect\citeauthoryear{Luo \bgroup et al\mbox.\egroup
  }{2018}]{luo2018knowledge}
Luo, K.; Lin, F.; Luo, X.; and Zhu, K.
\newblock 2018.
\newblock Knowledge base question answering via encoding of complex query
  graphs.
\newblock In {\em EMNLP},  2185--2194.

\bibitem[\protect\citeauthoryear{Manning \bgroup et al\mbox.\egroup
  }{2014}]{manning2014stanford}
Manning, C.; Surdeanu, M.; Bauer, J.; Finkel, J.; Bethard, S.; and McClosky, D.
\newblock 2014.
\newblock The stanford corenlp natural language processing toolkit.
\newblock In {\em ACL: System Demonstrations},  55--60.

\bibitem[\protect\citeauthoryear{Papineni \bgroup et al\mbox.\egroup
  }{2002}]{papineni2002bleu}
Papineni, K.; Roukos, S.; Ward, T.; and Zhu, W.-J.
\newblock 2002.
\newblock Bleu: a method for automatic evaluation of machine translation.
\newblock In {\em ACL},  311--318.

\bibitem[\protect\citeauthoryear{Pennington, Socher, and
  Manning}{2014}]{pennington2014glove}
Pennington, J.; Socher, R.; and Manning, C.
\newblock 2014.
\newblock Glove: Global vectors for word representation.
\newblock In {\em EMNLP},  1532--1543.

\bibitem[\protect\citeauthoryear{Serban \bgroup et al\mbox.\egroup
  }{2016}]{serban2015building}
Serban, I.~V.; Sordoni, A.; Bengio, Y.; Courville, A.; and Pineau, J.
\newblock 2016.
\newblock Building end-to-end dialogue systems using generative hierarchical
  neural network models.
\newblock In {\em AAAI},  3776--3783.

\bibitem[\protect\citeauthoryear{Song \bgroup et al\mbox.\egroup
  }{2018}]{song2018ensemble}
Song, Y.; Yan, R.; Li, C.-T.; Nie, J.-Y.; Zhang, M.; and Zhao, D.
\newblock 2018.
\newblock An ensemble of retrieval-based and generation-based human-computer
  conversation systems.
\newblock In {\em IJCAI},  4382--4388.

\bibitem[\protect\citeauthoryear{Sutskever, Vinyals, and
  Le}{2014}]{sutskever2014sequence}
Sutskever, I.; Vinyals, O.; and Le, Q.~V.
\newblock 2014.
\newblock Sequence to sequence learning with neural networks.
\newblock In {\em Advances in Neural Information Process-ing Systems},
  3104--3112.

\bibitem[\protect\citeauthoryear{Wu \bgroup et al\mbox.\egroup
  }{2018}]{wu2018response}
Wu, Y.; Wei, F.; Huang, S.; Li, Z.; and Zhou, M.
\newblock 2018.
\newblock Response generation by context-aware prototype editing.
\newblock {\em arXiv preprint arXiv:1806.07042}.

\bibitem[\protect\citeauthoryear{Xu \bgroup et al\mbox.\egroup
  }{2016}]{xu2016question}
Xu, K.; Reddy, S.; Feng, Y.; Huang, S.; and Zhao, D.
\newblock 2016.
\newblock Question answering on freebase via relation extraction and textual
  evidence.
\newblock In {\em ACL},  2326--2336.

\bibitem[\protect\citeauthoryear{Yao and Van~Durme}{2014}]{yao2014information}
Yao, X., and Van~Durme, B.
\newblock 2014.
\newblock Information extraction over structured data: Question answering with
  freebase.
\newblock In {\em ACL}, volume~1,  956--966.

\bibitem[\protect\citeauthoryear{Yih \bgroup et al\mbox.\egroup
  }{2015}]{yih2015semantic}
Yih, S. W.-t.; Chang, M.-W.; He, X.; and Gao, J.
\newblock 2015.
\newblock Semantic parsing via staged query graph generation: Question
  answering with knowledge base.
\newblock In {\em ACL},  1321--1331.

\bibitem[\protect\citeauthoryear{Yu \bgroup et al\mbox.\egroup
  }{2017}]{yu2017improved}
Yu, M.; Yin, W.; Hasan, K.~S.; Santos, C.~d.; Xiang, B.; and Zhou, B.
\newblock 2017.
\newblock Improved neural relation detection for knowledge base question
  answering.
\newblock In {\em ACL},  571--581.

\bibitem[\protect\citeauthoryear{Zhou \bgroup et al\mbox.\egroup
  }{2018}]{zhou2018commonsense}
Zhou, H.; Young, T.; Huang, M.; Zhao, H.; Xu, J.; and Zhu, X.
\newblock 2018.
\newblock Commonsense knowledge aware conversation generation with graph
  attention.
\newblock In {\em IJCAI},  4623--4629.

\bibitem[\protect\citeauthoryear{Zhu \bgroup et al\mbox.\egroup
  }{2017}]{zhu2017flexible}
Zhu, W.; Mo, K.; Zhang, Y.; Zhu, Z.; Peng, X.; and Yang, Q.
\newblock 2017.
\newblock Flexible end-to-end dialogue system for knowledge grounded
  conversation.
\newblock {\em arXiv preprint arXiv:1709.04264}.

\end{thebibliography}
\bibliographystyle{aaai}

\end{document}